\documentclass{article}

\usepackage[final]{corl_2018} 
\usepackage{amsmath}
\usepackage{amsthm}
\usepackage{amssymb}
\usepackage{color}
\usepackage{geometry}
\usepackage{graphicx}
\usepackage{caption}
\usepackage{subcaption}
\usepackage{float}

\graphicspath{{figures/}}

\title{Learning 6-DoF Grasping and \\ Pick-Place Using Attention Focus}

\author{
  Marcus Gualtieri\\
  College of Computer and Information Science\\
  Northeastern University,
  United States\\
  \texttt{mgualti@ccs.neu.edu} \\
  \And
  Robert Platt \\
  College of Computer and Information Science\\
  Northeastern University,
  United States\\
  \texttt{rplatt@ccs.neu.edu}\\
}

\begin{document}
\maketitle


\begin{abstract}
We address a class of manipulation problems where the robot perceives the scene with a depth sensor and can move its end effector in a space with six degrees of freedom -- 3D position and orientation. Our approach is to formulate the problem as a Markov decision process (MDP) with abstract yet generally applicable state and action representations. Finding a good solution to the MDP requires adding constraints on the allowed actions. We develop a specific set of constraints called hierarchical $\text{SE}(3)$ sampling (HSE3S) which causes the robot to learn a sequence of gazes to focus attention on the task-relevant parts of the scene. We demonstrate the effectiveness of our approach on three challenging pick-place tasks (with novel objects in clutter and nontrivial places) both in simulation and on a real robot, even though all training is done in simulation. 
\end{abstract}


\section{Introduction}




Reinforcement learning (RL) promises a way to train robots in dynamic environments with little supervision -- task is specified through a reward signal that could be changing or uninformative for a significant period of time. Unfortunately, classical RL does not work in realistic settings due to the high dimensional, continuous space of robot actions (position and orientation in 3D) and due to the complexity of the Markov state (all objects with their poses, velocities, shapes, masses, friction coefficients, etc.) Recent research addresses the complex state space by combining deep learning with RL and by representing state as a history of high resolution images \cite{Mnih2015}. However, the problem remains difficult due to noisy, incomplete sensor readings (partial observability), novel objects (generalization), and the high dimensional, continuous action space (curse of dimensionality). Additionally, deep RL methods tend to be data hungry which limits rapid, on-line learning.


Building on deep RL, we begin to address the issues of generalization and the complex action space by training a robot to focus on task-relevant parts of the scene. Restricting the robot's attention to task relevant areas is useful because incidental differences in scenes can be ignored and thus generalization is improved. Attention focus also narrows the space under which actions plausibly can be applied. In order to achieve this, our first step is to reformulate the problem as an MDP with abstract sense and motion actions. These abstractions induce a state space where it is at least possible for the robot to reason over only task-relevant information. Second, we develop a set of constraints on the sense actions which forces the robot to learn to focus on the task-relevant parts of the scene. This comes in the form of a hierarchical sequence of gazes that the robot must follow in order to focus attention on the task-relevant part of the action space. Third, we train in a simulated environment that can run many times faster than the real world and automatically provides information on task success. Direct transfer from simulation to the real world is possible due to the compact state representation and the way the robot learns to use it.


Specifically, contributions include (a) an abstract state-action representation to enable task-relevant focus (sense-move-effect MDP), (b) a set of constraints on the robot's action choices to enable learning of task-relevant focus (HSE3S), and (c) a single system which implements these ideas and can learn either 6-DoF grasping, 6-DoF placing, or joint 6-DoF grasping and placing by varying only the reward function. The effectiveness of the system is demonstrated both in simulation and on a physical UR5 robot on three challenging pick-place benchmark tasks. Although we focus on pick-place problems (the most fundamental manipulation primitives -- the robot must begin by grasping an object and end by placing it) many of the ideas are quite general, as discussed in Section~\ref{sec:conclusion}.

\section{Related Work}


Traditional approaches to robotic manipulation require accurate 3D models of the objects to be manipulated \cite{LozanoPerez1986}. Each object model has predetermined grasp and place poses associated with it, and the problem reduces to fitting the model to the sensor observation and calculating appropriate motion plans. Recent research has taken another direction -- grasping and placing novel objects, i.e., objects the robot has not been explicitly trained to handle. Most of these projects focus on grasping \cite{Lenz2015,Pinto2016,Mahler2017,tenPas2017,Zeng2018}, and the few that extend to pick-place (where the goal placement is non-trivial) assume either fixed grasp choices \cite{Jiang2012} or fixed place choices \cite{Gualtieri2018}. The present work generalizes these approaches by introducing a single system that can learn both grasping and placing.


Our approach to novel object manipulation leverages advances in deep RL. In particular, our learning agent is similar to that of deep Q-network (DQN) \cite{Mnih2015}; however, DQN has a simple, discrete list of actions, whereas our actions are continuous and high dimensional. Additionally, DQN  operates on the full images of the scene, whereas our robot learns to focus on the parts of the scene that are relevant for the task. More recent approaches to RL allow continuous, high-dimensional actions, in addition to operating on the full image of the scene \cite{Lillicrap2015,Levine2016A,Levine2016B}. These approaches take raw sensor observations (camera images) and output low-level robot actions (motor torques, velocities, or Cartesian motions). In contrast, state and action in our approach are more abstract -- state includes task-focused images and action includes target end effector pose -- in order to speed up learning and introduce robustness to transfer from a simulated domain to a real robot.  James, Davison, and Johns employ domain randomization to enable transfer of a policy learned in simulation to the real world \cite{James2017}, and Andrychowicz \textit{et al.} address the problem of exploration by replaying experiences with different goals \cite{Andrychowicz2017}. We view both of these improvements as complimentary: it is possible that our method could be improved by combining with these ideas.




Our strategy for sensing relies on learning a sequence of gazes in order to focus the robot's attention on the task-relevant part of the scene. This idea was inspired by work done with attention models and active sensing. Attention models are posed with the task of finding an object of interest in a 2D image \cite{Sprague2004,Larochelle2010,Mnih2014}. Our problem is much more difficult in that we select 6-DoF actions to maximize a generic reward signal. Active sensing is the problem of optimizing sensor placements for the achievement of some task \cite{Chen2011,Gualtieri2017}. However, we are mainly just interested in finding constraints to the sensing actions that allow efficient selection of end effector actions.

\section{Problem Statement}
\label{sec:problemStatement}

The state of the world consists of a robotic manipulator and several objects. Rigidly attached to the robot are an end effector and a depth sensor. The robot's configuration is completely described by a list of joint angles, $q = \left[q_1, \dots, q_n \right]$, $q_i \in \mathbb{R}$. The objects can have arbitrary poses and shapes. The robot's actions are delta motions of each joint, including the end effector joints: $\delta = \left[\delta_1, \dots, \delta_n \right]$, $\delta_i \in \mathbb{R}$. If there is no collision, $q_{t+1} = q_t + \delta_t$. On each time step the robot observes the environment through its depth sensor as a point cloud -- a list of 3D points in the reference frame of the sensor. The robot has a finite memory for storing up to the last $k$ observations.

The task is specified by a human operator or programmer via a reward function, e.g. $+1$ when an object is grasped and placed into a desired configuration and $0$ otherwise. The primary sources of uncertainty in the environment include (a) object surfaces are only partially visible in the current observation and (b) actions may not move the robot by the desired $\delta$ due to collisions. The goal of the learning algorithm is to learn a policy, i.e. a mapping from a history of observations to actions, that maximizes the expected sum of rewards per episode. The robot is not provided with a model of the environment -- it must learn through trial and error.  The robot is allowed to act for a maximum number of time steps per episode.

\section{Approach}
\label{sec:approach}


The first part of our approach is to reformulate the problem as an MDP with abstract states and actions (Section~\ref{sec:mdp}). This makes it possible for the robot to reduce the amount of sensor data it must process, assuming it takes the right actions. The second part of our approach is to add constraints to the allowed sense actions (Section~\ref{sec:hse3s}) which forces the robot to learn to focus attention on task-relevant parts of the scene. The problem is then amenable to solution via deep RL (Sections~\ref{sec:learningModel} and~\ref{sec:learningAlgorithm}).

\subsection{Sense-Move-Effect MDP}
\label{sec:mdp}

\begin{figure}[ht]
  \centering
  \includegraphics[height=0.75in]{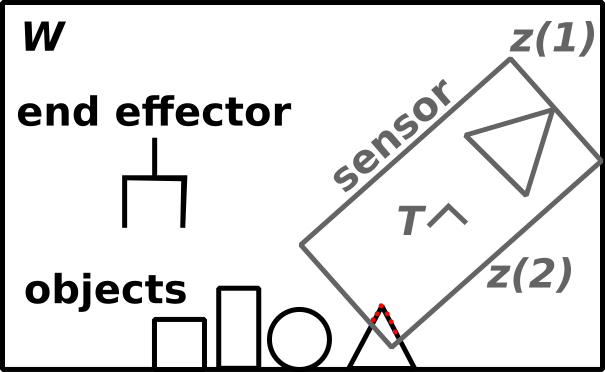}
  \includegraphics[height=0.75in]{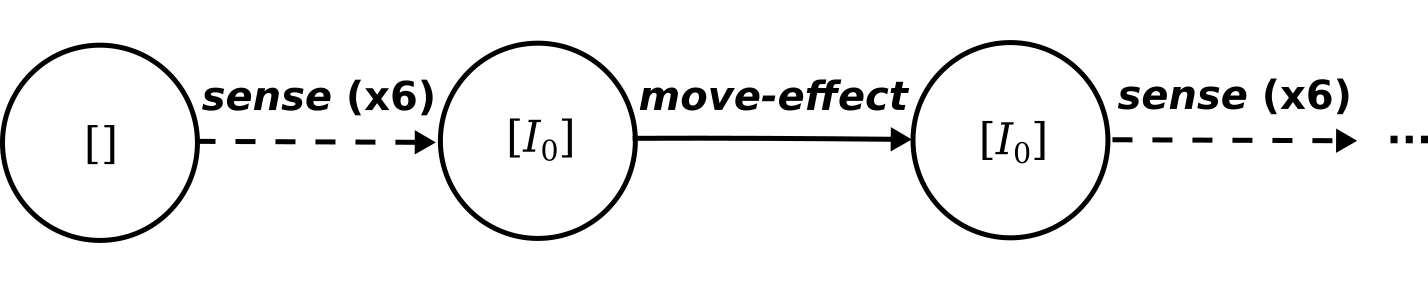}
  \caption{\textit{Left.} The abstract robot has a mobile sensor which observes points in a rectangular volume at pose $T$ with size $z$. The robot also has an end effector which can be moved to $T$ to perform some operation $o$. \textit{Right.} Sequence of actions the robot is allowed to take with state shown in circles.}
  \label{fig:simpleRobot}
  \label{fig:process}
\end{figure}

An MDP consists of states, actions, a transition model, and a reward function. The reward function is task specific, and we give examples in Sections~\ref{sec:simulation} and~\ref{sec:robot}. A diagram of the abstract robot is shown in Figure~\ref{fig:simpleRobot} (left).

\textbf{Action} consists of two high-level capabilities: \textit{sense}$(T, z)$ and \textit{move-effect}$(o)$. \textit{sense}$(T, z)$ uses the robot's sensor to acquire a point cloud $C$ with origin at $T \in W$, where $W \subseteq \text{SE}(3)$ is the reachable workspace of the robot's end effector, and where the point cloud is cropped to a rectangular volume of length, width, height $z \in \mathbb{R}^{+3}$. \textit{move-effect}$(o)$ moves the robot's end effector frame to $T$ and activates controller $o \in O$, where $O$ is a set of preprogrammed controllers actuating only end effector joints. As a running example, a binary gripper could be attached to the end effector with controllers $O = \{\textit{open}, \textit{close} \}$ providing capabilities for pick-place tasks.

\begin{figure}[ht]
  \centering
  \includegraphics[height=0.85in]{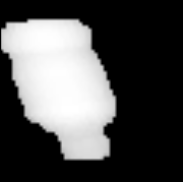}
  \includegraphics[height=0.85in]{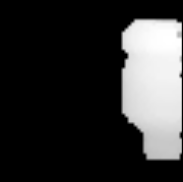}
  \includegraphics[height=0.85in]{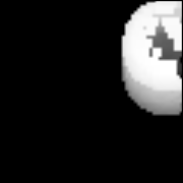}
  \includegraphics[height=0.85in]{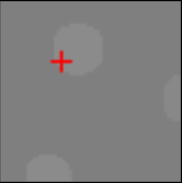}
  \includegraphics[height=0.85in]{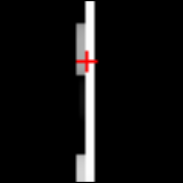}
  \includegraphics[height=0.85in]{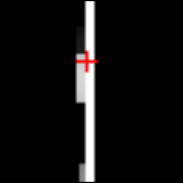}
  \caption{\textit{Left three.} Height maps of a the sensed point cloud of a bottle along z, y, and x directions just before \textit{move-effect(close)} was called. \textit{Right three.} Height maps of the currently sensed point cloud along z, y, and x directions. The red cross indicates the next position the robot intends to move to in order to place the bottle on a coaster.}
  \label{fig:stateImages}
\end{figure}

\textbf{State} consists of $k$ images, $(I_1, \dots, I_k)$, with $k \geq 1$. Each image has three channels corresponding to three, fixed-resolution height maps of $C$ as shown in Figure~\ref{fig:stateImages}. Since the height maps are of fixed resolution, larger values of $z$ will show more objects in the scene at less detail, and smaller values of $z$ will show only an object part in more detail. The purpose of using images instead of point clouds is to make the state encoding more compact and to allow it to work with modern convolutional neural network (CNN) architectures. To further reduce the size of the state representation, we only include the most recent observation -- because it represents the current location $T$ of the abstract sensor -- and only observations just before \textit{move-effect} actions -- because the environment changes only after \textit{move-effect} actions. For many pick-place tasks $k=2$ is sufficient because the two observations represent (a) the current target of the grasp/place and (b) the current hand contents. 


\textbf{Transition dynamics} are initially unknown to the robot and depend on the domain, e.g. simulation or real world. We assume a static environment except during \textit{move-effect} actions. In practice, \textit{move-effect} actions rely on a motion planner and controller to make the continuous motion from its current pose to $T$. These motions may fail due to planning failure or collisions.

\subsection{HSE3S}
\label{sec:hse3s}

The above MDP is still very difficult to solve if any action in the combined $W \times \mathbb{R}^{+3} \times O$ space is allowed. To make learning feasible, before each \textit{move-effect} action, the robot is constrained to follow a fixed number of \textit{sense} actions with predetermined values for $z$. This process is illustrated in Figure~\ref{fig:process} (right). We call this approach \textit{hierarchical SE$(3)$ sampling} (HSE3S). The basic idea is to select the next $sense$ position within the first, zoomed-out observation, decrease $z$ to focus on that point of interest, select another $sense$ position within this new observation, and so on for a fixed number of time steps, at which point the focused volume is small enough that few samples are needed to precisely choose the end effector pose.

In particular, the HSE3S approach specifies a list of tuples $\left[ (z_1, d_1) \dots, (z_{\bar t}, d_{\bar t}) \right]$, where $z_i$ is a predetermined $z$ value that the agent must use when taking \textit{sense}$(T, z)$ actions, and where $d_i \in \mathbb{R}^6$ is a predetermined allowed range of motion relative to the sensor's current location for choosing $T$. In other words, $z_i = [z(1), z(2), z(3)]$ sets the observation volume's length, width, and height for the $i$\textsuperscript{th} sense action in the sequence; and $d_i = [x, y, z, \theta, \phi, \rho]$ limits the allowed position offset to $\pm x$, $\pm y$, and $\pm z$ and rotational offset to $\pm \theta$, $\pm \phi$, and $\pm \rho$.

To see how this improves sample efficiency, imagine the problem of selecting position in a 3D volume. Let $\alpha$ be the largest volume allowed per sample. The number of samples $n_s$ needed is proportional to the initial volume $v_0 = z_0(1) z_0(2) z_0(3)$, i.e. $n_s= \lceil (1/\alpha)v_0 \rceil$. But with HSE3S, we select position sequentially, by say, halving the volume size in each direction at each time step, i.e. $z_{t+1} = 0.5 z_t$. In this case 8 samples are needed for each part of the divided space at each time step, so the sample complexity is $8\bar t$, where $\bar t$ is the total number of time steps, and $v_t = v_0/8^t$. To get $v_{\bar t} \leq \alpha$, $\bar t = \lceil \text{log}_8 (v_0 / \alpha) \rceil$, so the sample complexity is now logarithmic in $v_0$, but this is at the expense of having to learn to select the correct partition in each step.

We typically use HSE3S in situations where the observation is of a fixed resolution, due to limited sensing or processing capabilities, so that each consecutive \textit{sense} action reveals more detail of a smaller part of the scene. One potential issue with this partially observable case is that at the most zoomed-out level, objects may be blurry, and the robot may choose the wrong object without having a way to go back. Precisely, let \textit{sense-mislead} be when the highest valued \textit{sense} choice at time $t$ does not lead to the highest valued possible action at time $t+i$ for some positive $i$. To mitigate issues with sense-misleads, we can run HSE3S in $n$ independent trials, where the samples are different for each trial, and choose the trial that ends with the highest valued action choice. In practice, we take the 1-trial action during training when computational efficiency is more important than performance and the $n$-trial action during test time when performance is more important than efficiency. This is analogous to how Monte Carlo tree search is used on-line to improve policies \cite{Tesauro1997}.

\begin{figure}[ht]
  \centering
  \includegraphics[height=0.85in]{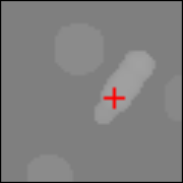}
  \includegraphics[height=0.85in]{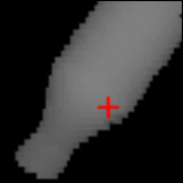}
  \includegraphics[height=0.85in]{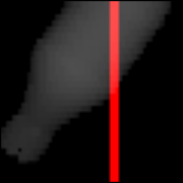}
  \includegraphics[height=0.85in]{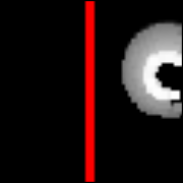}
  \includegraphics[height=0.85in]{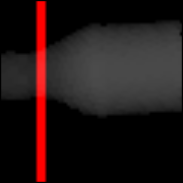}
  \includegraphics[height=0.85in]{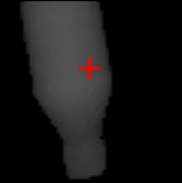}
  \caption{Example sense sequence for grasping a bottle: (a) coarse position, (b) fine position, (c) orientation about z, (d) orientation about y, (e) orientation about z, (f) final position. Crosses denote the position selected for the next action, and lines denote the orientation selected for the next action (where the center is 0 and edges $\pm \pi$). For all images besides (d) the gripper approaches the image plane with the fingers on the left and right sides.}
  \label{fig:hse3sSequence}
\end{figure}

The specific sense action sequence developed for our pick-place system is illustrated in Figures~\ref{fig:hse3sSequence}. The robot is initially presented with height maps of a point cloud centered in the robot's workspace and truncated at 36 cm$^3$. The robot then chooses a position, constrained to be on a point in this initial cloud. At the second level, the robot chooses any position in a 9 cm$^3$ volume, centered on the position selected in the previous level. At the third level, the robot chooses orientation about the current z-axis in $\pm \pi$ with $z=9 \text{ cm}^3$. At the fourth level, the robot chooses orientation about the current y-axis in $\pm \pi/2$ with $z = 9 \text{ cm}^3$. At the fifth level, orientation about z-axis in $\pm \pi$ with $z=9 \text{ cm}^3$. Finally, at the sixth level, position is chosen with $z = 10.5 \text{ cm}^3$, but this time the samples are $\pm 10.5$ cm only on the current x, y, or z axes. The purpose of this last level is to allow for a large offset when placing, such as placing on top of another object.

In addition to these the HSE3S constraints, we also constrain our pick-place system to follow a \textit{move-effect(close)}-\textit{move-effect(open)} sequence, since it does not make sense to open the gripper when it is already open and vis versa. We further ensure the gripper is in its open configuration at the beginning of each episode.

\subsection{Learning Model}
\label{sec:learningModel}

\begin{figure}[ht]
  \centering
  \includegraphics[height=2in]{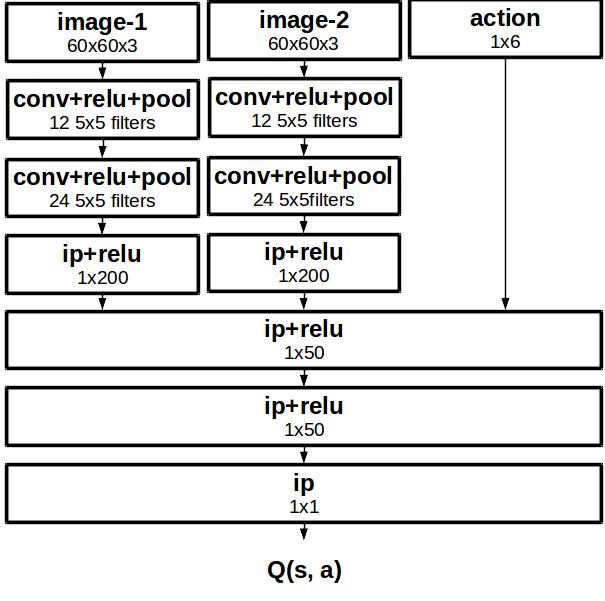}
  \includegraphics[height=2in]{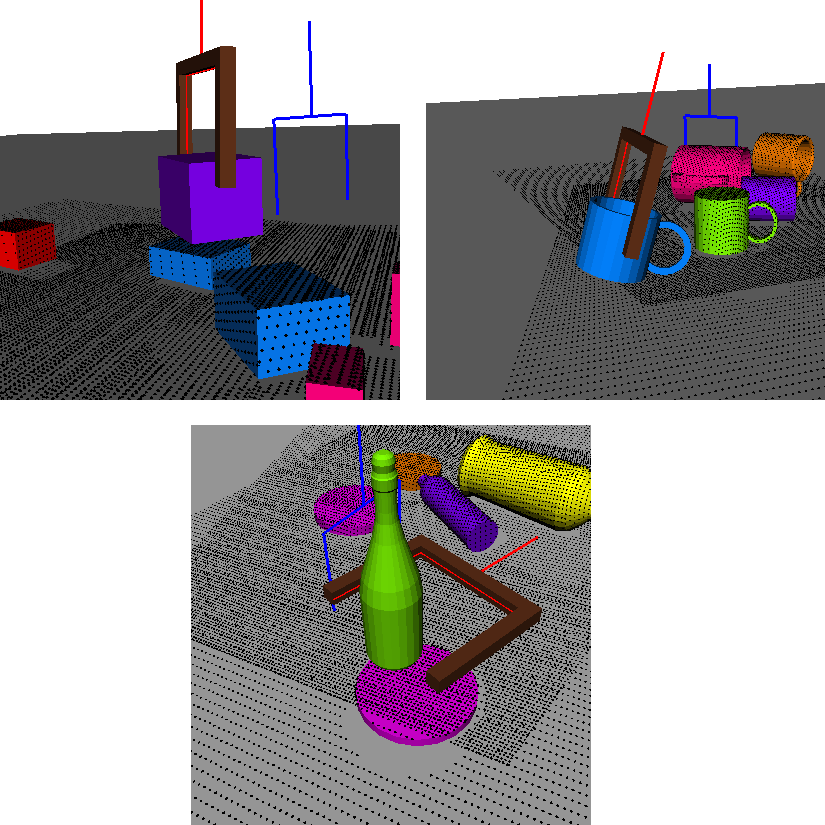}
  \caption{\textit{Left.} CNN architecture used to approximate $Q_t(s, a)$ for each step $t$ in a pick-place process. \textit{Right.} Example places for the three tasks in simulation. The blue marker shows from where the object was grasped. In each case, all required and partial credit conditions were met, except with the bottle, where it was placed too high for the partial credit height condition.}
  \label{fig:cnn}
  \label{fig:simTasks}
\end{figure}

For each time step in our MDP, the allowed actions and image zoom levels are different. Thus, we factor the Q-function into several independent CNNs, one for each time step. The architecture for each is the same and is illustrated in Figure~\ref{fig:cnn} (left). Any unused input is set to zero; e.g., when selecting orientation, one element of the action vector is set to an  Euler angle and the other five are zero.

\subsection{Learning Algorithm}
\label{sec:learningAlgorithm}

Our learning algorithm is similar to DQN \cite{Mnih2015} except instead of Q-learning we use gradient Monte Carlo \cite{Sutton2018} for the update rule. That is, for each state-action pair $(s_t, a_t)$ in an experience replay buffer, we assign the sum of rewards experienced from that time forward $r_t + r_{t+1} + \cdots + r_{\bar t}$ as the training label. We empirically found Monte Carlo to perform just as well as Sarsa \cite{Rummery1994} for a grasping task (\textit{c.f.} \cite{Quillen2018}), and it is computationally more efficient due to not having to evaluate the Q-function when generating training labels.

\section{Simulation Experiments}
\label{sec:simulation}


We evaluated our approach on three pick-place tasks in simulation as depicted in Figure~\ref{fig:simTasks}. For the first task, the robot is presented with 2 to 10 blocks cluttered together. The goal is to grasp one block and stack it on top of another block. For the second task, the robot is presented with 1 to 5 mugs in clutter. The goal is to grasp one mug and place it upright on the table. In the third task the robot is presented with 1 to 3 bottles and 3 round coasters. The goal is to grasp a bottle and place it upright on a coaster. We used OpenRAVE for the simulation environment \cite{Diankov2010} and 3DNet for non-primitive objects (i.e. bottles and mugs) \cite{Wohlkinger2012}.  The robot is given no prior information on how to grasp or place other than the constraints mentioned in Section~\ref{sec:approach} -- the robot must learn these skills from trial and error.

Ideally, the reward function for each scenario should simply describe the set of goal states, e.g. +1 when the object is placed correctly and $0$ otherwise. However, successful 6-DoF pick-places are rare at the beginning of learning when actions are random. To make the reward signal more informative, we assign a reward in $[0,1]$ for a successful grasp and an additional reward in $[0,1]$ for a successful place. For both grasping and placing there are required conditions and partial credit conditions. All required conditions must be met for the reward to be nonzero; otherwise the reward is proportional to the number of partial credit conditions met. For example, when placing bottles, the required conditions include: (a) all required grasp conditions, (b) the bottle is within $30^\circ$ of upright, (c) above a coaster, (d) no more than 4 cm above the coaster, and (c) not in collision or not in collision if moved up 2 cm. The partial credit conditions include: (a) within $15^\circ$ of upright, (b) no more than 2 cm above the coaster, and (d) not in collision. Similar reward functions were developed for both grasping and placing stages for the other tasks.


\begin{figure}[ht]
  \centering
  \includegraphics[height=1.35in]{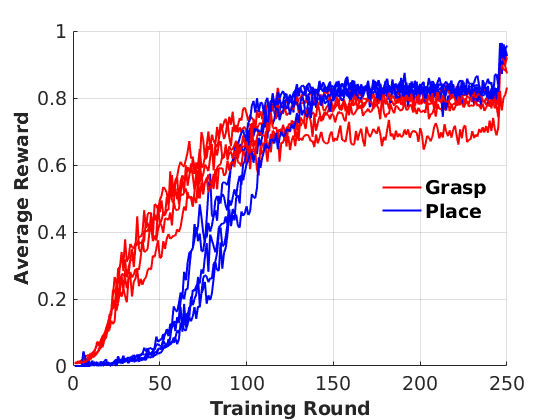}
  \includegraphics[height=1.35in]{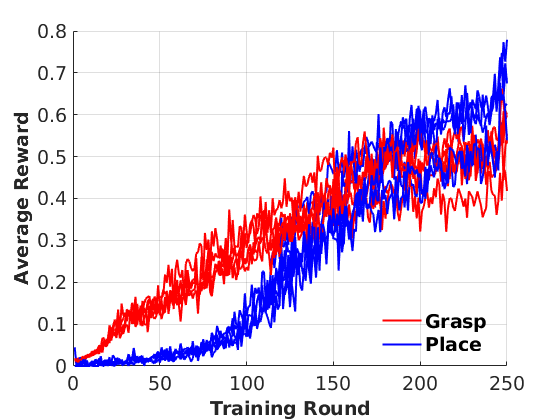}
  \includegraphics[height=1.35in]{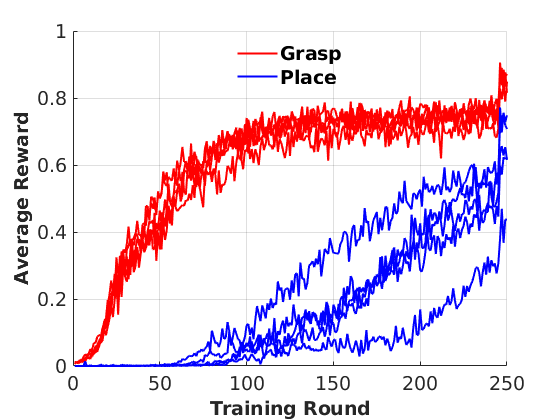}
  \caption{\textit{Left.} Blocks. \textit{Center.} Mugs. \textit{Right.} Bottles. Grasp and place rewards shown in different colors -- episode return is the sum of the two. Rewards are averages over all of the episodes in a training round. Each training round consisted of $2,000$ episodes followed by $3,000$ iterations of stochastic gradient descent. Maximum possible reward for a grasp/place is 1. Each plot includes learning curves for five independent runs. All episodes used $1$-trial sampling (Section \ref{sec:hse3s}).}
  \label{fig:learningCurves}
\end{figure}

Learning curves for the three benchmark tasks are shown in Figure~\ref{fig:learningCurves}. The exploration strategy was $\epsilon$-greedy, where $\epsilon=1$ at the beginning (i.e. completely random actions), gradually drops down to $0.05$ by the 25\textsuperscript{th} training round for grasping ($\epsilon$-place drops as more place attempts are experienced due to successful grasps), and $\epsilon=0$ for the last 5 training rounds. In terms of task difficulty, placing blocks appears easiest, followed by placing bottles on coasters, followed by placing mugs on the table. This ordering is consistent with the ambiguity of the object pose given a partial point cloud: blocks have three planes of symmetry, bottles have two orthogonal planes of symmetry, and mugs have at most one plane of symmetry (an upright plane dividing the handle). Also, there is a high variance between training realizations placing bottles on coasters, and this in turn may be due to the longer time it takes to learn the first several successful placements. We suspect this is because the space of goal placements is small compared to the blocks scenario (where the desired orientation is less specific) or the mugs scenario (where anywhere on the table is fine).


\begin{table}[ht]
  \centering
  \begin{tabular}{|c|c|c|c|c|c|c|} 
  \hline
   & Blocks A & Blocks A$\land$CF & Mugs A & Mugs A$\land$CF & Bottles A & Bottles A$\land$CF\\
  \hline
  GPD & 0.95 & 0.51 & 0.77 & 0.52 & \bf 0.98 & 0.71\\
  \hline
  $1$-trial & 0.86 & 0.82 & 0.65 & 0.58 & 0.83 & 0.81\\
  \hline
  $10$-trial & \bf 0.99 & \bf 0.98 & \bf 0.85 & \bf 0.80 & 0.94 & \bf 0.93\\
  \hline
  \end{tabular}
  \caption{Grasp success rates averaged over $1,000$ independent episodes. Performance is shown with the antipodal condition \cite{tenPas2017} only (A) and with both antipodal and collision-free conditions (A$\land$CF). The comparison is with GPD \cite{tenPas2017}, 1-trial HSE3S, and 10-trial HSE3S. GPD used 500 grasp samples and the highest-scoring grasp for evaluation. Both GPD and HSE3S were trained with 3DNet objects in the named category.}
  \label{tab:gpd}
  \vspace{-0.1in}
\end{table}

For the three pick-place tasks, grasping and placing are learned jointly. Unlike related work where mechanically stable grasping is an assumed capability \cite{Jiang2012,Gualtieri2018}, a 6-DoF grasp pose is selected by the robot to maximize the expected reward for the full pick-place task. However, our same system can be easily modified to learn mechanically stable grasping independently of placing. We compared the grasping performance of our system to an off-the-shelf grasping software, grasp pose detection (GPD) \cite{tenPas2017}. The result is summarized in Table~\ref{tab:gpd}. For blocks, the HSE3S approach gets perfect grasps about 98\% of the time whereas GPD gets perfect grasps only 51\% of the time. Both systems were trained on the same set of blocks; however, the evaluation was done in the same environment (e.g. same simulated depth sensor) that the HSE3S system was trained in, so it is not surprising HSE3S does better. However, the evaluation for stable grasps is nearly the same for both methods (antipodal, force closure grasps \cite{tenPas2017}), and GPD uses a much larger representation for the state images (includes surface normals). The collision criteria in simulation may be overly conservative, so we also show success rates with the antipodal condition alone.


In the above analysis with grasping, we used the $n$-trial sampling strategy to minimize sense-mislead errors as described in Section~\ref{sec:hse3s}. The HSE3S sequence was run 10 independent times with random sampling, and the highest valued state-action at the end was taken for \textit{move-effect(close)}. Without $n$-trial sampling, HSE3S gets perfect grasps only 82\% of the time for blocks. We would expect similar performance gains at the end of the learning curves in Figure~\ref{fig:learningCurves} by $n$-trial sampling.

\section{Robot Experiments}
\label{sec:robot}

\begin{figure}[ht]
  \centering
  \includegraphics[height=0.9in]{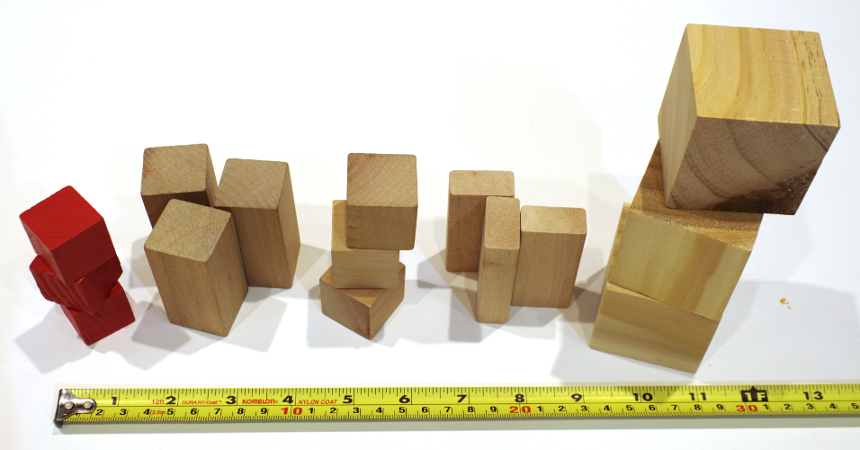}
  \includegraphics[height=0.9in]{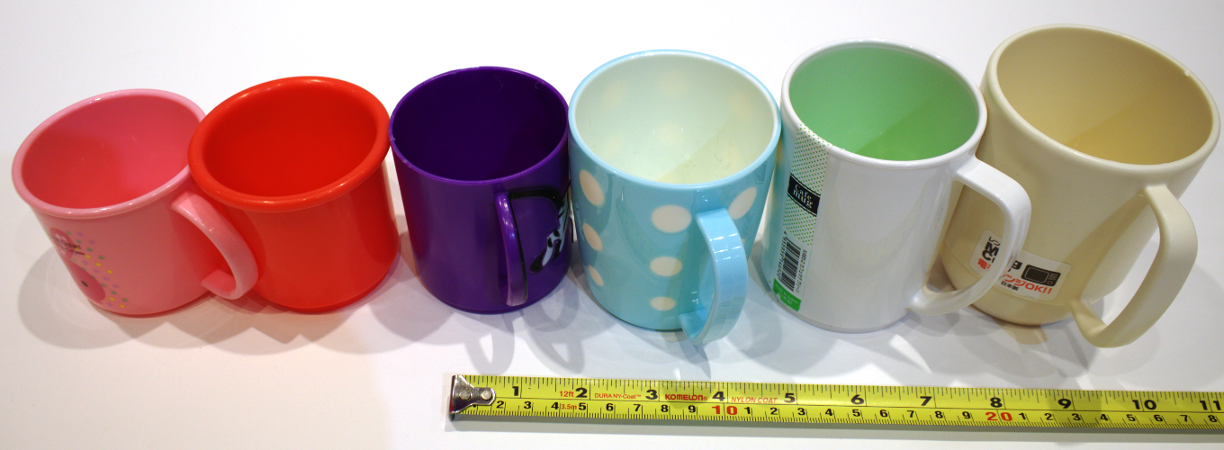}
  \includegraphics[height=0.9in]{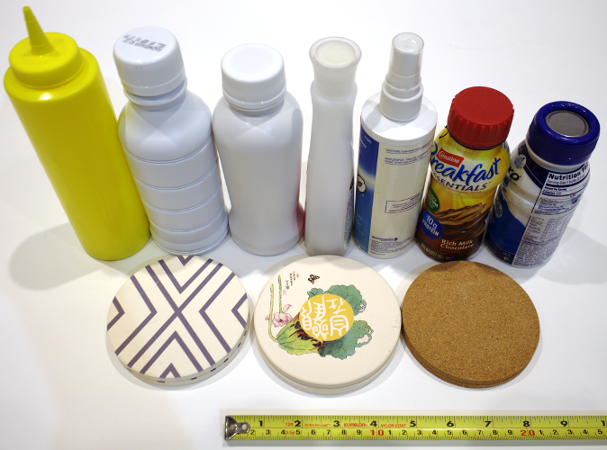}
  \caption{\textit{Left.} Blocks test set. 10/15 blocks were randomly selected for each round. \textit{Center.} Mugs test set. 5/6 mugs were randomly selected for each round. \textit{Right.} Bottles test set. 3/7 bottles were randomly selected for each round. All 3 coasters were present in each round.}
  \label{fig:testSet}
\end{figure}

We tested our approach on a UR5 robot equipped with a Robotiq 85 parallel-jaw gripper and a wrist-mounted Structure depth sensor. The policies learned in simulation were directly transfered to the physical system without additional training. Objects were selected to be diverse, able to fit in the robot's gripper, and visible to the sensor. The complete test sets are shown in Figure~\ref{fig:testSet}. Prior to each episode, the robot acquired two point clouds approximately $45^\circ$ above the table on opposite sides. This point cloud was transformed, cropped, and projected into images as the robot followed the HSE3S sequence. That is, in our experiments the \textit{sense} actions operate virtually on a point cloud acquired from fixed sensor positions; however, in principle, for each \textit{sense} action, the robot could physically move the sensor to acquire a point cloud centered on the target \textit{sense} volume.

\begin{figure}[ht]
  \centering
  \includegraphics[trim={0in 0.5in 0in 0.0in},clip,height=1.22in]{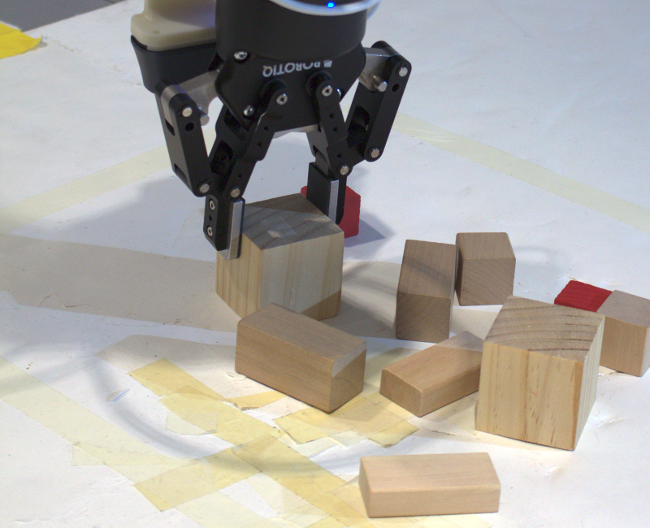}
  \includegraphics[trim={0in 0.5in 0in 0.0in},clip,height=1.22in]{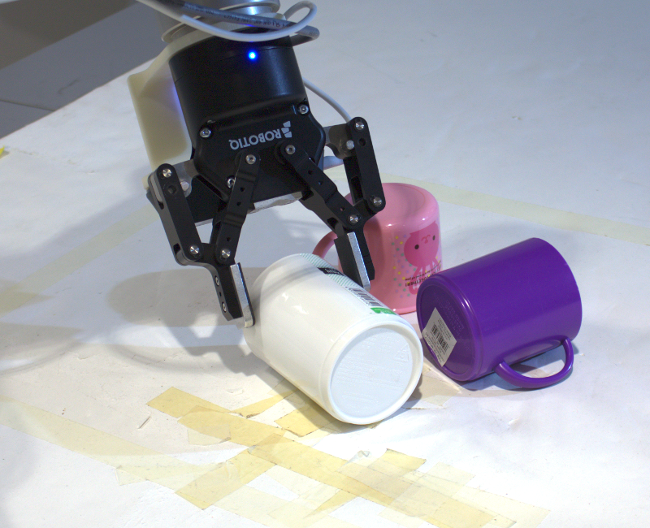}
  \includegraphics[trim={0in 0.0in 0in 0.5in},clip,height=1.22in]{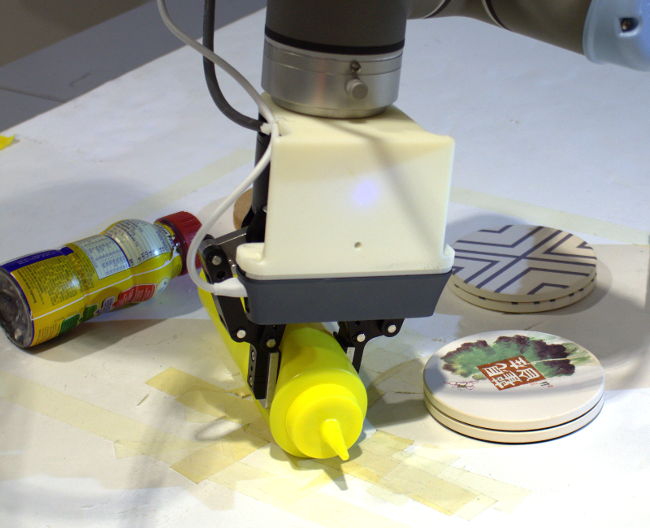}\\
  \includegraphics[trim={0in 0.5in 0in 0.0in},clip,height=1.22in]{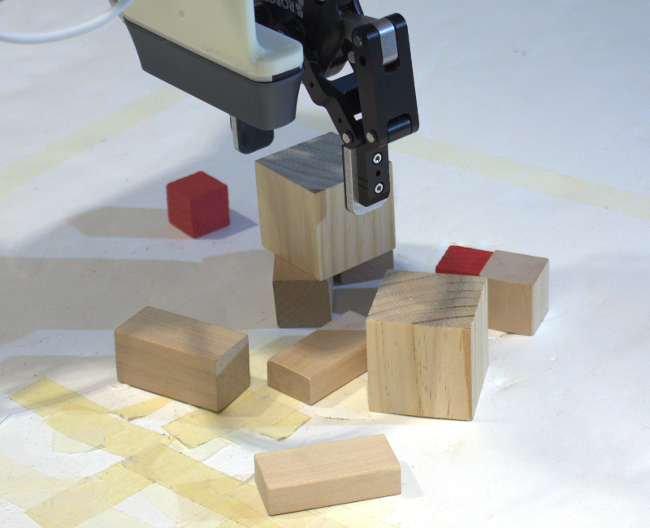}
  \includegraphics[trim={0in 0.5in 0in 0.0in},clip,height=1.22in]{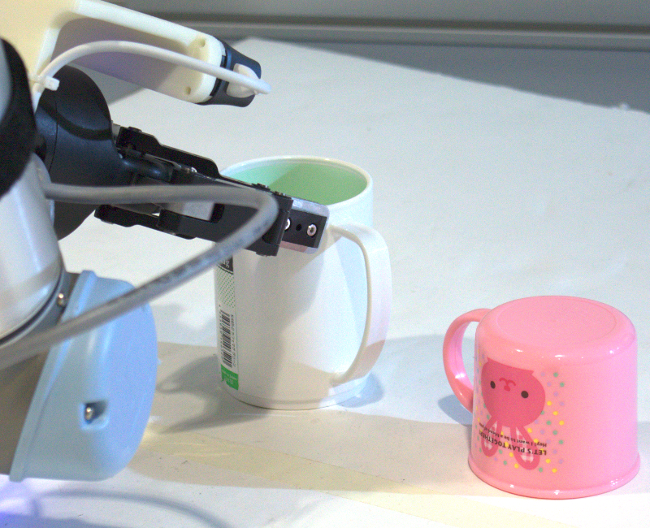}
  \includegraphics[trim={0in 0.5in 0in 0.0in},clip,height=1.22in]{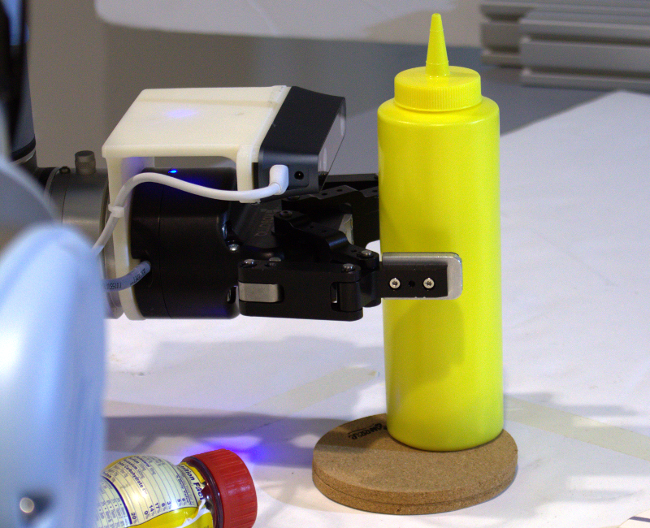}
  \caption{\textit{Left.} Blocks task: one block must be placed on another. \textit{Center.} Mugs task: one mug must be placed upright on the table. Notice the grasp is diagonal to the mug axis, and the robot compensates for this by placing diagonally with respect to the table surface. \textit{Right.} Bottles task: place a bottle on a coaster.}
  \label{fig:robot}
\end{figure}

The tasks were much the same as in simulation. For the first task, 10 blocks were randomly selected from the test set and dumped onto the table. The robot's goal was to place one block on top of another. After each place or each failure, the block interacted with by the robot was removed by a human. This continued until only 1 block remained. For the second task, 5 mugs were selected from the test set and dumped onto the table, and the robot was to place them upright on the table. After each place or failure, the mug interacted with was removed. This continued until no mugs remained. For the third task, the 3 bottles and 3 coasters were placed arbitrarily except no bottle was allowed to touch any coaster. Again, after each place or failure, a bottle was removed, and this continued until no bottles remained. Examples of these tasks being attempted by the robot are shown in Figure~\ref{fig:robot}.

In the real robot experiments, $n$-trial sampling, described in Section~\ref{sec:hse3s}, played an additional role besides mitigating issues with sense-misleads. Several (about 100) samples were evaluated for grasps and places, rejected at each level if values were below a threshold, and sorted by the last value in the sense sequence. Then IK and motion planning was attempted for each grasp/place pose, starting with the highest valued pose, and continuing down the list until a reachable pose was found. For the mug and bottle tasks, additional place poses were added at the original place position and rotated about the gravity axis. This degree of freedom in the object reference frame should not affect place success, and this helped with finding reachable, high valued place poses.


Table~\ref{tab:robot} shows the grasp, place, and task success rates for each task. Results are shown with and without detection failures. A \textit{detection failure} occurred when all grasp or place values were below a threshold. In these cases, the robot believes there is no good option available in the scene and, in a deployed scenario, would ask for human assistance. The major failure for block placing was misalignments with the bottom block. We believe this is primarily due to incorrect point cloud registration with the robot's base frame. Precise point cloud alignment was important because many of the place surfaces were small -- the smallest block size was 2 cm$^3$. For mugs, grasp failures were the main problem. This corresponds to simulation results where mug grasp performance was lower than in the other two scenarios (Figure~\ref{fig:learningCurves}). For bottles, there were a combination of problems, mainly placing upside-down, diagonally, or too close to the edge of the coaster. The poorer place performance for bottles also corresponds with simulation results.

\begin{table}[ht]
  \centering
  \begin{tabular}{|c|c|c|c|c|c|c|} 
  \hline
  & Blocks (w/ d.f.) & Blocks & Mugs (w/ d.f.) & Mugs & Bottles (w/ d.f.) & Bottles\\
  \hline
  Grasp & 0.96 & 0.96 & 0.86 & 0.86 & 0.89 & 0.90\\
  \hline
  Place & 0.67 & 0.67 & 0.89 & 0.91 & 0.64 & 0.67\\
  \hline
  Task & 0.64 & 0.64 & 0.76 & 0.78 & 0.57 & 0.60\\
  \hline
  \hline
  $n$ Grasps & 50 & 50 & 51 & 50 & 53 & 50\\
  \hline
  $n$ Places & 48 & 48 & 44 & 43 & 47 & 45\\
  \hline
  \hline
  DF & 0 & N/A & 1 & N/A & 3 & N/A\\
  \hline
  GF & 2 & 2 & 7 & 7 & 5 & 5\\
  \hline
  FOS & 13 & 13 & N/A & N/A & 6 & 6\\
  \hline
  PUD & N/A & N/A & 4 & 4 & 7 & 7\\
  \hline
  PIS & 3 & 3 & 0 & 0 & 2 & 2\\
  \hline
  \end{tabular}
  \caption{\textit{Top.} Success rates for blocks, mugs, and bottles tasks, separately including detection failures (w/ d.f.) and not counting attempts where there was a detection failure. \textit{Middle.} Number of grasp and place attempts. A place was attempted only after the grasp was successful. \textit{Bottom.} Failure counts broken down by type: detection failure (DF), grasp failure (GF), fell off support surface (FOS), placed upside-down (PUD), and placed into support surface (PIS).}
  \label{tab:robot}
  \vspace{-0.2in}
\end{table}

\section{Conclusion}
\label{sec:conclusion}


Although we focus on pick-place tasks with simple grippers, the ideas developed in this paper could be extended to other tasks. HSE3S could be used to improve sample efficiency of any method that samples poses in SE$(3)$, at the cost of having to learn where to gaze. Also, the deictic \cite{Whitehead1991} state representation has been used in this and previous work \cite{Gualtieri2018} with much empirical success at transferring a policy learned in simulation to the real world. Also, one may imagine additional tasks that could be performed with the present system if more primitive actions were available, such as opening a drawer may require a \textit{move-in-line}$(v)$ action that moves from the current pose along vector $v$.


For future work, we plan to compare HSE3S to actor critic methods like DDPG \cite{Lillicrap2015} where action is output from the function approximator. Outputting the action directly avoids the need to sample actions; however, sampling actions has the benefit of inducing a nondeterministic policy useful in situations with reachability constraints. We also plan to examine how this approach scales to broader object categories, e.g. kitchen items. Finally, we would would like to demonstrate the approach on problems with longer time horizons. For instance, the three benchmark tasks could become block stacking, placing all mugs upright, and placing all bottles on coasters. Such problems may require an even more compact state representation -- since state would consist of a longer history -- and further speedups to learning -- since informative rewards would be delayed.



\clearpage
\acknowledgments{We thank Andreas ten Pas for many helpful discussions while this project was underway and the reviewers for their valuable feedback. This work has been supported in part by the NSF through IIS-1427081, IIS-1724191, and IIS-1724257; NASA through NNX16AC48A and NNX13AQ85G; ONR through N000141410047; Amazon through an ARA to Platt; and Google through an FRA to Platt.}


\bibliography{References}  

\end{document}